\documentclass[conference]{IEEEtran}
\IEEEoverridecommandlockouts
\usepackage{cite}
\usepackage{amsmath,amssymb,amsfonts}
\usepackage{algorithmic}
\usepackage{graphicx}
\graphicspath{{./Figures/}}
\usepackage{textcomp}
\usepackage{xcolor}

\usepackage[boxruled, linesnumbered]{algorithm2e}
\usepackage[utf8]{inputenc}
\usepackage{hyperref,ragged2e}
\usepackage{enumitem}
\usepackage{subcaption}
\usepackage{array}
\usepackage{multirow}
\usepackage{footnote,tablefootnote}
\usepackage{xcolor}

\def\BibTeX{{\rm B\kern-.05em{\sc i\kern-.025em b}\kern-.08em
    T\kern-.1667em\lower.7ex\hbox{E}\kern-.125emX}}
\begin{document}

\title{Training Deep Neural Networks with Constrained Learning Parameters\\
}

\makeatletter
    \newcommand{\linebreakand}{%
      \end{@IEEEauthorhalign}
      \hfill\mbox{}\par
      \mbox{}\hfill\begin{@IEEEauthorhalign}
    }
\makeatother

\author{\IEEEauthorblockN{Prasanna Date}
\IEEEauthorblockA{\textit{Department of Computer Science} \\
\textit{Rensselaer Polytechnic Institute}\\
Troy, New York 12180 \\
datep@rpi.edu}
\and
\IEEEauthorblockN{Christopher D. Carothers}
\IEEEauthorblockA{\textit{Department of Computer Science} \\
\textit{Rensselaer Polytechnic Institute}\\
Troy, New York 12180 \\
chris.carothers@gmail.com}
\and
\IEEEauthorblockN{John E. Mitchell}
\IEEEauthorblockA{\textit{Department of Mathematical Sciences} \\
\textit{Rensselaer Polytechnic Institute}\\
Troy, New York 12180 \\
mitchj@rpi.edu}
\linebreakand
\IEEEauthorblockN{James A. Hendler}
\IEEEauthorblockA{\textit{Department of Computer Science} \\
\textit{Rensselaer Polytechnic Institute}\\
Troy, New York 12180 \\
hendler@cs.rpi.edu}
\and
\IEEEauthorblockN{Malik Magdon-Ismail}
\IEEEauthorblockA{\textit{Department of Computer Science} \\
\textit{Rensselaer Polytechnic Institute}\\
Troy, New York 12180 \\
magdon@cs.rpi.edu}
}

\maketitle

\begin{abstract}
Today's deep learning models are primarily trained on CPUs and GPUs.
Although these models tend to have low error, they consume high power and utilize large amount of memory owing to double precision floating point learning parameters.
Beyond the Moore's law, a significant portion of deep learning tasks would run on edge computing systems, which will form an indispensable part of the entire computation fabric.
Subsequently, training deep learning models for such systems will have to be tailored and adopted to generate models that have the following desirable characteristics: \emph{low error}, \emph{low memory}, and \emph{low power}.
We believe that deep neural networks (DNNs), where learning parameters are constrained to have a set of finite discrete values, running on neuromorphic computing systems would be instrumental for intelligent edge computing systems having these desirable characteristics.
To this extent, we propose the Combinatorial Neural Network Training Algorithm (CoNNTrA), that leverages a coordinate gradient descent-based approach for training deep learning models with finite discrete learning parameters.
Next, we elaborate on the theoretical underpinnings and evaluate the computational complexity of CoNNTrA.
As a proof of concept, we use CoNNTrA to train deep learning models with ternary learning parameters on the MNIST, Iris and ImageNet data sets and compare their performance to the same models trained using Backpropagation.
We use following performance metrics for the comparison: (i) Training error; (ii) Validation error; (iii) Memory usage; and (iv) Training time.
Our results indicate that CoNNTrA models use $32\times$ less memory and have errors at par with the Backpropagation models.
\end{abstract}

\begin{IEEEkeywords}
Deep Neural Networks, Training Algorithm, Deep Learning, Machine Learning, Artificial Intelligence
\end{IEEEkeywords}

\section{Introduction}
\label{sec:intro}
Deep neural networks (DNNs) have had a significant impact on our lives in the twenty first century---from advancing scientific discovery \cite{karpatne2017theory} to improving the quality of life \cite{sharma2019iot,sundaravadivel2018smart,naylor2018prospects}.
DNNs are trained using traditional learning algorithms like Backpropagation on conventional computing platforms using CPUs and GPUs.
While DNNs trained using this approach have low error and can be trained in a reasonable amount of time, they consume large amount of memory and power.
This will not be sustainable in the post Moore's law era, where a significant portion of deep learning tasks will be ported to edge computing systems \cite{aimone2019neural}.
Edge computing systems would form an indispensable part of the entire computation fabric and support critical applications like Internet of Things (IoT), autonomous vehicles, embedded systems etc. \cite{mailhiot2018energy}.
Therefore, it is important to train DNNs that are tailored for such systems and have the following three desirable characteristics: \emph{low error}, \emph{low memory}, and \emph{low power}.

While low power could be achieved using neuromorphic computing systems \cite{schuman2017survey}, we focus on achieving low error and low memory in this work.
Our work in this paper can potentially be extended to neuromorphic systems to achieve these desirable characteristics.
In order to achieve low memory, we focus on deep learning models where learning parameters are constrained to have a set of finite discrete values, for example, binary or ternary values.
By constraining the values of learning parameters, we significantly reduce the memory required to store them.
For instance, a learning parameter constrained to have ternary values ($-1$, $0$, $+1$) can be stored using just $2$ bits, as opposed to a double precision floating point learning parameter used in traditional learning algorithms, which requires $64$ bits.

To train DNNs with constrained learning parameters, we propose a novel training algorithm called the Combinatorial Neural Network Training Algorithm (CoNNTrA) in Section \ref{sec:conntra}.
Our objective is to demonstrate that CoNNTrA can train deep learning models consuming significantly less memory, yet achieving errors at par with Backpropagation.
In Section \ref{sec:performance-evaluation}, we use CoNNTrA to train deep learning models for three machine learning benchmark data sets (MNIST, Iris and ImageNet).
We compare the performance of CoNNTrA to that of Backpropagation along four performance metrics: training error, validation error, memory usage and training time.
Our results indicate that CoNNTrA models have errors at par with Backpropagation, and consume $32\times$ less memory.

\section{Related Work}
\label{sec:related}
Deep learning models with binary learning parameters have been proposed in the literature for several use cases.
Courbariaux et al. propose BinaryConnect, which can train binary neural networks for specialized hardware and test their approach on MNIST, CIFAR-10 and SVHN data sets \cite{courbariaux2015binaryconnect}.
Rastegari et al. propose XNOR-Net, which can train binary convolutional neural networks (CNN), test their algorithm on the ImageNet data set, and report $32\times$ savings in memory and $58\times$ faster convolutional operations \cite{rastegari2016xnor}.
Wan et al. propose Ternary Binary Network (TBN), having ternary inputs and binary learning parameters, for edge computing devices like portable devices and wearable devices, test their approach on ImageNet and PASCAL VOC data sets and achieve $32\times$ memory savings and $40\times$ faster convolutional operations \cite{wan2018tbn}.
Andri et al. propose YodaNN, a hardware accelerator for BinaryConnect CNNs and obtain high power efficiency \cite{andri2016yodann}.

In addition to binary and ternary neural networks, several approaches have been proposed in the literature to train quantized neural networks.
Hubara et al. propose a method to train quantized neural networks having low precision weights and test their approach on the MNIST, CIFAR-10, SVHN and ImageNet data sets \cite{hubara2017quantized}.
Zhou et al. propose a mechanism for iterative optimizations for training quantized neural network and test their approach on AlexNet, GoogLeNet and ResNet \cite{zhou2017iqnn}.
Blott et al. describe an end-to-end deep learning framework for exploration and training of quantized neural networks that can optimize for a given platform, design target or specific precision \cite{blott2018finn}.
Choi et al. propose a mechanism for parameterized clipping activation for quantized neural networks that enables training with low precision weights \cite{choi2018pact}.

We have previously shown that training deep neural networks with constrained learning parameters is an NP-complete problem \cite{date2019combinatorial}.
To address this problem, several evolutionary optimization-based approaches have been pursued in the literature.
Shen et al. propose an evolutionary optimization-based learning mechanism that finds binary neural networks by searching through the entire search space of learning parameters \cite{shen2019searching}.
Too et al. use a binary particle swarm optimization for feature extraction and compare their approach to other evolutionary optimization-based approaches that leverage genetic algorithm, binary gravitational search algorithm and competitive binary grey wolf optimizer \cite{too2019new}.
Nogami et al. use a combination of genetic algorithm and simulated annealing to optimize the bin boundaries of quantization for CNN and test their approach on the ImageNet data set using AlexNet and VGG16 \cite{nogami2019optimizing}.

There are several limitations of the approaches proposed in the literature, especially with regards to low error, low memory and low power edge computing systems.
While the algorithms to train binary or ternary neural networks have the potential to be deployed on edge computing systems, they are very specialized and cannot be used to train neural networks with a set of finite discrete learning parameters directly.
On the other hand, while evolutionary optimization-based methods produce accurate models for quantized neural networks, they cannot be deployed on edge computing systems because evolutionary optimization is a compute heavy process and may require large compute clusters.
Moreover, most of the approaches proposed in the literature cater to a specific deep learning model, for example, convolutional neural network and it is unclear if they are useful for other deep learning models such as recurrent neural networks or generative adversarial networks.

In this work we propose the Combinatorial Neural Network Training Algorithm (CoNNTrA), which is not restricted to any particular neural network architecture, has an efficient time complexity (polynomial), and does not necessarily require significant amount of compute power.
CoNNTrA is not restricted to binary or ternary learning parameters specifically, but can train any configuration of learning parameters as long as they are finite and discrete.
We believe CoNNTrA would be able to train deep neural networks having low error, low memory and low power for edge computing systems in the post Moore's law era, especially when combined with neuromorphic computing systems, which are known to be resilient and energy efficient \cite{schuman2020resilience}, and have a wide range of applications such as graph algorithms \cite{hamilton2020spike,kay2020neuromorphic}, modeling epidemics \cite{hamilton2020modeling} and predicting supercomputer failures \cite{date2018efficient}.





\section{The DNN Training Problem}
\label{sec:snn-training-problem}

We define the DNN training problem using the following notation:
\begin{itemize}
    \item $\mathbb{R}$, $\mathbb{N}$, $\mathbb{B}$: Set of real numbers, natural numbers and binary numbers ($\mathbb{B} = \{0,1\}$) respectively.
    \item $\mathbb{T}$: Ternary set $\mathbb{T} = \{-1, 0, +1\}$.
    \item $\omega$: Set of finite discrete values that the learning parameters $W$ can take, for example, if learning parameters are required to have binary values, $\omega = \mathbb{B}$.
    \item $N$: Number of points in the training dataset.
    \item $d$: Dimension of each point in training dataset, which is the same as number of features in the training dataset.
    \item $k$: Number of classes for classification.
    \item $X$: $X$ can be a scalar, vector, matrix or tensor containing training data.
    \item $Y$: $Y$ contains the labels of training data encoded in a one-hot format. Since we have $N$ data points and $k$ classes, $Y \in \mathbb{B}^{N \times k}$.
    \item $W$: Set of all learning parameters, including all the weights and biases.
    \item $g(X, W)$: The DNN learning function.
    \item $e(P, Y)$: The error function which computes the error between predicted labels $P$ and ground truth labels $Y$.
\end{itemize}

Given training data $X$ and training labels $Y$, we would like to learn the parameters $W$ of the learning function $g(X, W)$ by minimizing the error $e(P, Y)$. 
In this regard, the DNN training problem with finite discrete weights is defined as follows:
\begin{align}
    \underset{W}{\min} \quad e(P, Y) \label{eq:snn-training}
\end{align}
where, 
$P = g(X, W)$ are the labels predicted by the learning function $g$;
Each learning parameter in the set $W$ can take values from the finite discrete set $\omega$.

\section{DNN Training with Constrained Learning Parameters is NP-Hard}
\label{sec:np-hard}

We show that under the Euclidean error function, training a single layer neural network with binary weights is NP-Hard by reducing the quadratic unconstrained binary optimization (QUBO) problem, which is known to be NP-Hard \cite{wang2009analyzing,boros2007local,pardalos1992complexity}, to the DNN training problem with finite discrete weights.
We first define the QUBO problem:
\begin{align}
    \text{The QUBO Problem:} \quad \min_{z \in \mathbb{B}^d} z^T A z + z^T b + c \label{eq:qubo}
\end{align}
where,
$A$ is a real symmetric positive definite $d \times d$ matrix, $b$ is a $d$-dimensional vector, and $c$ is a real scalar.
Note that if $A$ is not symmetric, it can be made symmetric by setting $a_{ij} = \frac{a_{ij} + a_{ji}}{2} \quad \forall i \ne j$, without changing the QUBO problem.
It is known that even when $A$ is positive definite, the QUBO problem is NP-Hard \cite{kochenberger2014unconstrained}.

The DNN training problem with binary weights and Euclidean error function is defined as follows:
\begin{align}
    \min_{W \in \mathbb{B}^d} e(P,Y) = \frac{1}{N} || P - Y ||^2_2 \label{eq:binary-snn-training}
\end{align}
where,
$P = g(X, W)$ is the vector of values predicted by the learning function $g(X, W) = X^T W$, $X \in \mathbb{R}^{N \times d}$, $Y \in \mathbb{R}^N$, $W \in \mathbb{B}^d$.
After expanding Equation \ref{eq:binary-snn-training}, the SNN Training problem becomes:
\begin{align}
    \min_{W \in \mathbb{B}^d} \frac{1}{N} (W^T X^T X W - 2 W^T X^T Y + Y^T Y) \label{eq:binary-snn-training-expanded}
\end{align}

Given the optimal solution, we can compute the objective function value in polynomial time, so the problem is in NP.
Also, Equation \ref{eq:binary-snn-training-expanded} is very similar to Equation \ref{eq:qubo}, in that both are quadratic minimization problems with binary variables.
In order to reduce the QUBO problem to Binary SNN Training, we first decompose the real symmetric positive definite QUBO matrix $A$ into a product of a unique lower triangular matrix with real positive diagonal entries $L$ and its transpose $L^T$ using the Cholesky decomposition:
\begin{align}
    A \xrightarrow[]{\text{CHOLESKY}} L L^T
\end{align}

Because $L$ is a lower triangular matrix with real positive diagonal entries, $L^{-1}$ exists.
The reduction is performed as follows:
\begin{itemize}
\item $W = z$
\item $\frac{X^T X}{N} = A = L L^T $ \\
        Therefore, $X = \sqrt{N} L^T $ and $X^T = \sqrt{N} L $
\item $\frac{-2 X^T Y}{N} = b$ \\
        Therefore, $Y = -\frac{\sqrt{N}}{2} L^{-1} b$
\item Because both QUBO and Binary SNN Training are unconstrained optimization problems, the scalars $c$ in QUBO and $\frac{1}{N}Y^T Y$ in Binary SNN Training do not affect the optimal solution. 
In order to equate the scalars in both problems, we can introduce another scalar $c'$ in Binary SNN Training so that $\frac{Y^T Y + c'}{N} = c$ without changing the optimal solution.

\end{itemize}

By setting $W = z$, $X = \sqrt{N} L^T$ and $Y = -\frac{\sqrt{N}}{2} L^{-1} b$, we have reduced the QUBO problem to Binary SNN Training problem, thus showing that Binary SNN Training problem is NP-Hard.
With more complex error functions like softmax and complex neural network architectures like deep convolutional or recurrent neural networks, the SNN training problem is at least as hard as the QUBO problem, if not more.

\section{Combinatorial Neural Network Training Algorithm (CoNNTrA)}
\label{sec:conntra}

\begin{algorithm}[t!]
	\caption{Discretization Subroutine for CoNNTrA}
	\label{algo:discretize}
	\SetAlgoLined
	\SetKwProg{Fn}{Function}{:}{}
	\SetKwFunction{discretize}{Discretize}
	\SetKwFunction{sort}{Sort}
	\SetKwFunction{zeros}{Zeros}

	\Fn{\discretize{$W_{pre}$, $\omega$}}{
		\KwIn{\\
			$W_{pre}$: Pretrained Weights \\
			$\omega$: Set of Finite Discrete Values
		} 
		
		\BlankLine
		\BlankLine

		\KwOut{ \\
			$W$: Discretized Weights
		}

		\BlankLine
		\BlankLine

		Weights: $W =$ \zeros{$|W_{pre}|$}\;
	    $\omega =$ \sort{$\omega$}\;
	    \For{$i = 1$ \KwTo $|W_{pre}|$}{
	        \For{$j = 1$ \KwTo $|\omega|$}{
	            \If{ $j == |\omega|$}{
	                $W[i] = \omega[j]$;
	            }
	            \ElseIf{$W_{pre}[i] \le \frac{1}{2} (\omega[j] + \omega[j+1])$}{
	                $W[i] = \omega[j]$\;
	                \textbf{break}\;
	            }
	        }
	    }

	    \BlankLine
	    \BlankLine

	    \Return $W$    
	}
\end{algorithm}

\begin{algorithm}[t!]
    \caption{CoNNTrA: Combinatorial Neural Network Training Algorithm}
    \label{algo:CoNNTrA}
    \SetAlgoLined
    \SetKwProg{Fn}{Function}{:}{}
    \SetKwFunction{conntra}{CoNNTrA}
    \SetKwFunction{randint}{RandomInteger}
    \SetKwFunction{discretize}{Discretize}

    \Fn{\conntra{$X$, $Y$, $W_{pre}$, $\omega$, $g(X,W)$, $e(P, Y)$}}{
        \KwIn{\\
            $X$: Training Data \\ 
            $Y \in \mathbb{B}^{N \times k}$: Training Labels (One-Hot Format) \\
            $W_{pre}$: Pretrained Weights (Reshaped into a Single 1-Dimensional Array) \\
            $\omega$: Set of Finite Discrete Values \\
            $g(X, W)$: Spiking Neural Network Function \\
            $e(P,Y)$: Error Function
            }
    
        \BlankLine
        \BlankLine
        
        \KwOut{ \\
            $W_{opt}$: Optimal Weights \\
            $\epsilon_{opt}$: Optimal Error
            }
    
    \BlankLine
    \BlankLine
    
    \tcc{PHASE 1: DISCRETIZATION}
    Weights: $W = $ \discretize{$W_{pre}$, $\omega$}

    \BlankLine
    \BlankLine

    \tcc{PHASE 2: INITIALIZATION}
    Error: $\epsilon = e(g(X, W), Y)$\;
    Optimal Weights: $W_{opt} = W$\;
    Optimal Error: $\epsilon_{opt} = \epsilon$\;
    Number of training iterations: $T$\;

    \BlankLine
    \BlankLine
    
    \tcc{PHASE 3: TRAINING}
    \For{$t = 1$ \KwTo $T$}{
        \For{$i = 1$ \KwTo $|W|$}{
            $i^{'} =$ \randint{$|W|$}\; 
            \For{$j = 1$ \KwTo $|\omega|$}{
                $W[i^{'}] = \omega[j]$\;
                $\epsilon = e(g(X, W), Y)$\;
                \If{$\epsilon \le \epsilon_{opt}$}{
                    $\epsilon_{opt} = \epsilon$\;
                    $W_{opt} = W$\;
                }
            }
            $W[i^{'}] = W_{opt}[i^{'}]$\;
        }
    }
    
    \BlankLine
    \BlankLine
    
    \Return $W_{opt}, \epsilon_{opt}$
}
\end{algorithm}

We propose the Combinatorial Neural Network Training Algorithm (CoNNTrA), which is a coordinate gradient descent based algorithm for training DNNs with finite discrete weights.
CoNNTrA is presented in Algorithm \ref{algo:discretize} and Algorithm \ref{algo:CoNNTrA}.
Say $w_i = \omega_j$ for some $i$ and $j$.
We first look at the left and right gradients of the error function.
\begin{align}
    & \text{Left Gradient:} \nonumber \\ 
    & \qquad \frac{\partial e}{\partial w_i} \Bigg|_{\text{left}} = \lim_{h \rightarrow 0} \frac{e(w_i) - e(w_i - h)}{h} \\
    & \text{Right Gradient:} \nonumber \\ 
    & \qquad \frac{\partial e}{\partial w_i} \Bigg|_{\text{right}} = \lim_{h \rightarrow 0} \frac{e(w_i + h) - e(w_i)}{h}
\end{align}

These gradients could be used if $w_i$ could take continuous values.
Since $w_i$ cannot take continuous values, $h$ never tends to $0$ in the above equations, but is some finite number greater than $0$.
So, we look at the discrete counterparts of gradients:
\begin{align}
    &\text{Left Discrete Gradient:} \nonumber \\
    &\qquad \frac{\Delta e}{\Delta w_i} \Bigg|_{\text{left}} = \frac{e(w_i = \omega_j) - e(w_i = \omega_{j-1})}{\omega_j - \omega_{j-1}} \\
    &\text{Right Discrete Gradient:} \nonumber \\
    &\qquad \frac{\Delta e}{\Delta w_i} \Bigg|_{\text{right}} = \frac{e(w_i = \omega_{j+1}) - e(w_i = \omega_j)}{\omega_{j+1} - \omega_j} 
\end{align}

These discrete counterparts of gradients search in the discrete vicinity of $w_i$ to find a value that lowers the error.
This is a local search, and makes up to three calls to the error function, i.e. $e(w_i = \omega_{j-1})$, $e(w_i = \omega_j)$ and $e(w_i = \omega_{j+1})$.
We extend this notion of local search and do a global search, i.e. search through all possible values of $w_i$ to find the best value that minimizes the error function.
This makes $\mathcal{O}(|\omega|)$ calls to the error function.
In this case, we have a better chance of finding a lower value of error function at each iteration.
When we do a similar procedure for all the weights, we iteratively find better weight values that decrease the error function gradually as training progresses.

CoNNTrA takes as inputs the training data $X$, the training labels $Y$, pretrained weights $W$, set of finite discrete values that the weights can take $\omega$, the SNN function $g(X, W)$ and the error function $e(P, Y)$.
Initial weights are the weights obtained when the DNN was trained using Backpropagation by relaxing the finite discrete value constraint.
The first step is to discretize the weights using Algorithm \ref{algo:discretize}.
In Algorithm \ref{algo:discretize}, we set all the weights in the vicinity of $\omega_j$ to $\omega_j$.
For example, if $\omega = \{-1, 0, +1\}$, then for some $i$, if the pretrained weight $w_i > 0.5$, it would be set to $+1$, if $-0.5 < w_i \le 0.5$, it would be set to $0$, and if $w_i \le -0.5$, it would be set to $-1$.
We start by initializing the weights $W$ to an array of zeros using the function \texttt{Zeros($x$)}, which returns a zero initialized array of length $x$, in line 2 of Algorithm \ref{algo:discretize}.
Next, we sort $\omega$ so that all values in $\omega$ are in increasing order in line 3.
Next, in the for loop from line 4 through 14, we iterate over each weight in $W$, and in the for loop from line 5 through 13, we iterate over each value in $\omega$ to find an appropriate discretized value for each weight.
The discretized weight value is assigned to the appropriate weight on either line 7 or 10.

In the second phase of the algorithm, i.e. the initialization phase, we first compute the error using the discretized weights $W$ and assign it to the initial error $\epsilon$ on line 3.
Next, we initialize the optimal weights $W_{opt}$ and optimal error $\epsilon_{opt}$, by setting them to $W$ and $\epsilon$ on lines 4 and 5 respectively.
We then define the number of training iterations $T$.

In the third phase of the algorithm, i.e. the training phase, we iterate over $T$ training iterations in lines 7--20.
During each iteration, we perform a global search over $|W|$ randomly selected weights in lines 8--19.
We refer to each random selection of weights as an epoch---so, there are a total of $T \times |W|$ training epochs.
During each training epoch, we first randomly select a weight index $i^{'}$ using the function \texttt{RandomInteger($x$)}, which returns a uniform random integer in the interval $[1, x]$.
For $w_{i^{'}}$, we perform a global search over all possible values of $w_{i^{'}}$ to find the best value that minimizes the error function in lines 10--17 of Algorithm \ref{algo:CoNNTrA}.
If a better value for $w_{i^{'}}$ is found, we update the optimal error $\epsilon_{opt}$ and optimal weights $W_{opt}$ in lines 14 and 15 respectively.
After a global search is performed for $w_{i^{'}}$, we set the current weights $W$ to the optimal weights $W_{opt}$ in line 18, so that in the subsequent epochs, we use the current best set of weights.
Finally, after all the training epochs are completed, we return $W_{opt}$ and $\epsilon_{opt}$ in line 21 of Algorithm \ref{algo:CoNNTrA}.

\subsection{Time Complexity}
\label{sub:time-complexity}
We analyze the running time of CoNNTrA by going over the running time of each line in Algorithm \ref{algo:discretize} and Algorithm \ref{algo:CoNNTrA}.
In the first phase, initializing the weights (line 2 of Algorithm \ref{algo:discretize}) takes $\mathcal{O}(|W|)$ time, and sorting $\omega$ takes $\mathcal{O}(|\omega| \log |\omega|)$ time.
Next, the two for loops in lines 4 through 14 of Algorithm \ref{algo:discretize} take up $\mathcal{O}(|W| \cdot |\omega|)$ time.
So the running time of discretization phase is $\mathcal{O}(|W| \cdot |\omega|)$.
We assume that time taken to do a forward pass on the SNN (i.e. computing $P = g(X, W)$) and computing the error $e(P, Y)$ takes $\tau = \mathcal{O}(e(g(X, W), Y))$ amount of time.
Therefore, it takes $\mathcal{O}(\tau)$ time to compute the error on line 3 of Algorithm \ref{algo:CoNNTrA}.
It takes $\mathcal{O}(|W|)$ time to initialize $W_{opt}$ on line 4 of Algorithm \ref{algo:CoNNTrA}.
Lines 5 and 6 take $\mathcal{O}(1)$ time.
So, the initialization phase takes $\mathcal{O}(\tau + |W|)$ time.
In the training phase, the for loop from lines 7 through 20 in Algorithm \ref{algo:CoNNTrA} runs $T$ times.
The for loop from lines 8 through 19 runs $|W|$ times and the for loop from lines 10 through 17 runs $|\omega|$ times.
It takes $\mathcal{O}(\tau)$ time to compute the error $\epsilon$ on line 14.
Therefore, the training phase takes $\mathcal{O}(T \cdot |W| \cdot |\omega| \cdot \tau)$ time.
Since this dominates the running time of all phases, the running time for CoNNTrA is $\mathcal{O}(T \cdot |W| \cdot |\omega| \cdot \tau)$.
Since $\tau$ is usually a polynomial time expression in the number of weights and size of training dataset, CoNNTrA is a polynomial time algorithm.

\subsection{Convergence}
\label{sub:convergence}
During each epoch in the training phase, we update the optimal weights $W_{opt}$ and optimal error $\epsilon_{opt}$ only if the current error $\epsilon$ is lower than $\epsilon_{opt}$.
Thus, with every update, $\epsilon_{opt}$ gets closer and closer to 0.0 demonstrating convergence.
If CoNNTrA is run for enough number of epochs, the optimal error would converge to a local minimum.

\section{Performance Evaluation}
\label{sec:performance-evaluation}

We compare the performance of CoNNTrA (Algorithm \ref{algo:CoNNTrA}) to traditional Backpropagation using GPU on four benchmark problems: MNIST using a logistic regression classifier, MNIST using a convolutional neural network (CNN), Iris using a deep neural network (DNN), and ImageNet using a convolutional neural network (CNN).
The performance metrics used for this comparison are:
\begin{enumerate}[topsep=0pt, partopsep=0pt, itemsep=0pt, parsep=0pt]
    \item Training Error: Percentage of data points classified incorrectly in the training dataset.
    \item Validation Error: Percentage of data points classified incorrectly in the validation dataset.
    \item Memory Usage (kilobytes): Amount of memory used to store the weights.
    \item Training Time (seconds): Total time taken to complete training.
\end{enumerate}

CoNNTrA was written in Python using the Numpy library \cite{van2011numpy}.
The Backpropagation algorithm was run using the TensorFlow library \cite{abadi2016tensorflow} on GPUs.
All experimental runs were run on a machine that had 32 cores of two-way multi-threaded Intel Xeon CPUs running at 2.60 GHz, three NVIDIA GPUs (GeForce GTX 1080 Titan, GeForce GTX 950 and GeForce GTX 670), 112 GB DIMM Synchronous RAM, 32 KB L1 cache, 256 KB L2 cache and 20 MB L3 cache.

\subsubsection{MNIST Logistic Regression}



\begin{figure}[t!]
	\centering
	\includegraphics[trim={100 0 100 0},clip,scale=0.3]{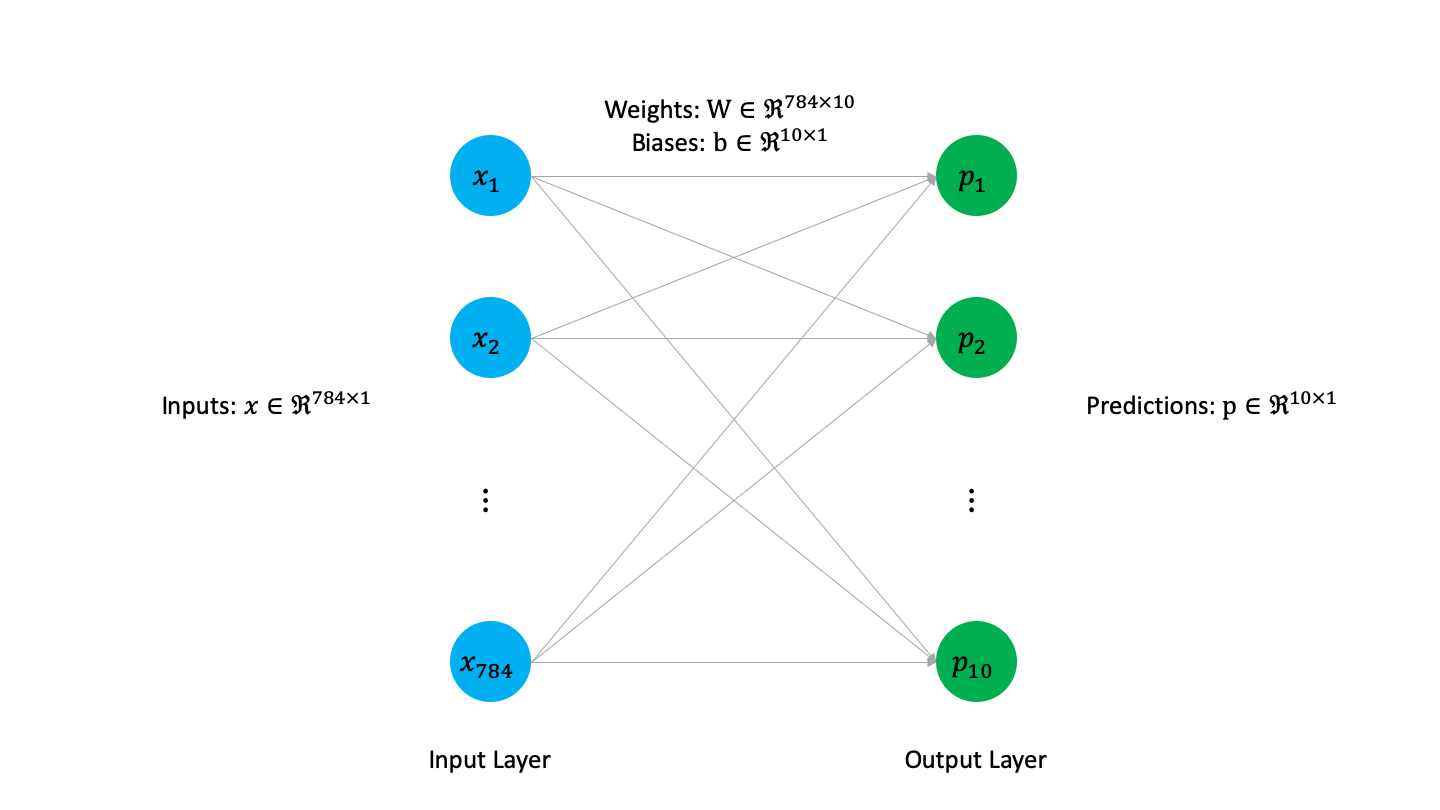}
	\caption{Schematic diagram of MNIST logistic regression model}
	\label{fig:mnist-logreg-model}
\end{figure}

\begin{table}[t!]
    \centering
    \caption{Performance metrics for MNIST logistic regression}
    \begin{tabular}{m{0.2\textwidth} m{0.11\textwidth} m{0.07\textwidth}}
        \noalign{\smallskip} \hline \noalign{\smallskip}
        Performance Metric & Backpropagation & CoNNTrA \\
        \noalign{\smallskip} \hline \noalign{\smallskip}
        Training Error (\%) 		          & 6.25		        & 7.59 \\
        Validation Error (\%) 		          & 7.34		        & 8.44 \\
        \textbf{Memory Usage (kilobytes)} 	  & \textbf{62.8}		& \textbf{1.96} \\
        Training Time (seconds) 	          & 81.70 	            & 236.12 \\
        \noalign{\smallskip} \hline \noalign{\smallskip} 
    \end{tabular}
    \label{tab:mnist-logreg-comparison}
\end{table}

\begin{figure}[t!]
    \centering
    \begin{subfigure}{0.4\textwidth}
        \centering 
        \includegraphics[scale=0.4]{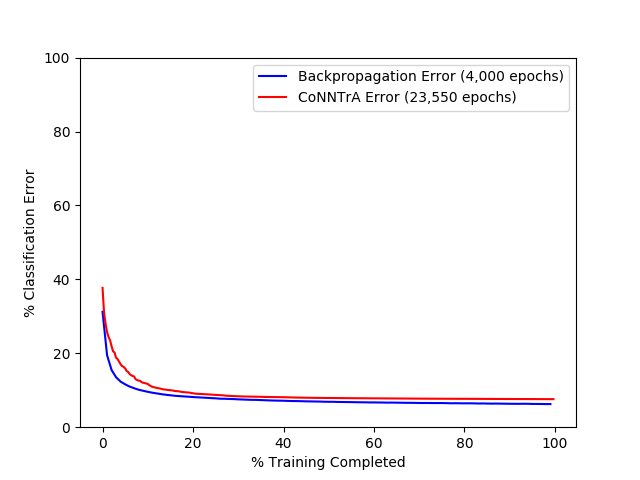}
        \caption{Training Error Comparison}
        \label{fig:mnist-logreg-training-error}
    \end{subfigure}
    \begin{subfigure}{0.4\textwidth}
        \centering 
        \includegraphics[scale=0.4]{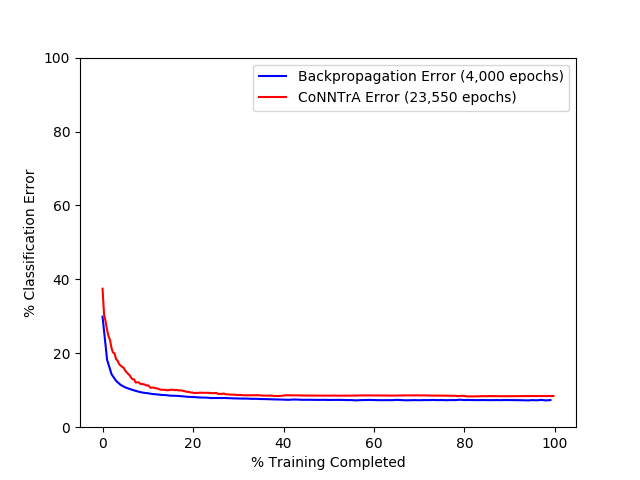}
        \caption{Validation Error Comparison}
        \label{fig:mnist-logreg-validation-error}
    \end{subfigure}
    \caption{Error comparison for MNIST logistic regression models}
    \label{fig:mnist-logreg-errors}
\end{figure}

We used a logistic regression model to classify the MNIST images.
The inputs to the logistic regression model were vectorized MNIST images, each of size $784 \times 1$.
The outputs to the model were the labels of the input images encoded in a one-hot format.
The model consisted of a weight matrix of size $784 \times 10$ and a bias vector of size $10 \times 1$.
A schematic diagram of the logistic regression model is shown in Figure \ref{fig:mnist-logreg-model}.
The activation function for this model was softmax and the loss was computed using the cross entropy loss function.

Table \ref{tab:mnist-logreg-comparison} shows the performance metrics of Backpropagation and CoNNTrA for the MNIST task using a logistic regression classifier.
The training errors for Backpropagation and CoNNTrA are $6.25\%$ and $7.59\%$ respectively, and the validation errors are $7.32\%$ and $8.44\%$ respectively.
The memory usage for Backpropagation and CoNNTrA is $62.8$ and $1.96$ kilobytes respectively.
While Backpropagation takes $81.70$ seconds to complete training, CoNNTrA takes $236.12$ seconds.
Figure \ref{fig:mnist-logreg-errors} shows the plot of training and validation errors for CoNNTrA (red) and Backpropagation (blue).
These errors were computed as percentage of misclassified points in the training and validation datasets respectively.
The X-axis in Figures \ref{fig:mnist-logreg-training-error} and \ref{fig:mnist-logreg-validation-error} shows the percentage of training completed.
The Y-axis shows the classification errors as a percentage.
As training progresses, both algorithms converge to the same ballpark of $6-8\%$, which corresponds to an accuracy of $92-94\%$.

\subsubsection{MNIST CNN}

\begin{table}[t!]
    \centering
    \caption{Performance metrics for MNIST CNN}
    \begin{tabular}{m{0.2\textwidth} m{0.11\textwidth} m{0.07\textwidth}}
        \noalign{\smallskip} \hline \noalign{\smallskip}
        Performance Metric & Backpropagation & CoNNTrA \\
        \noalign{\smallskip} \hline \noalign{\smallskip}
        Training Error (\%) 		        & 1.40		         & 2.60 \\
        Validation Error (\%) 		        & 1.56 		         & 2.39 \\
        \textbf{Memory Usage (kilobytes)} 	& \textbf{649.55}	 & \textbf{20.30} \\
        Training Time (seconds) 	        & 121.97	         & 4,871.04 \\
        \noalign{\smallskip} \hline \noalign{\smallskip} 
    \end{tabular}
    \label{tab:mnist-cnn-comparison}
\end{table}

\begin{figure}[t!]
    \centering
    \begin{subfigure}{0.4\textwidth}
        \centering 
        \includegraphics[scale=0.4]{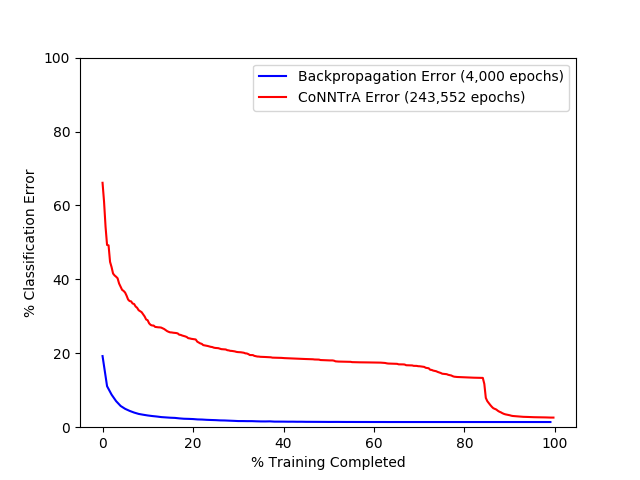}
        \caption{Training Error Comparison (MNIST CNN)}
        \label{fig:mnist-cnn-training-error}
    \end{subfigure}
    \begin{subfigure}{0.4\textwidth}
        \centering 
        \includegraphics[scale=0.4]{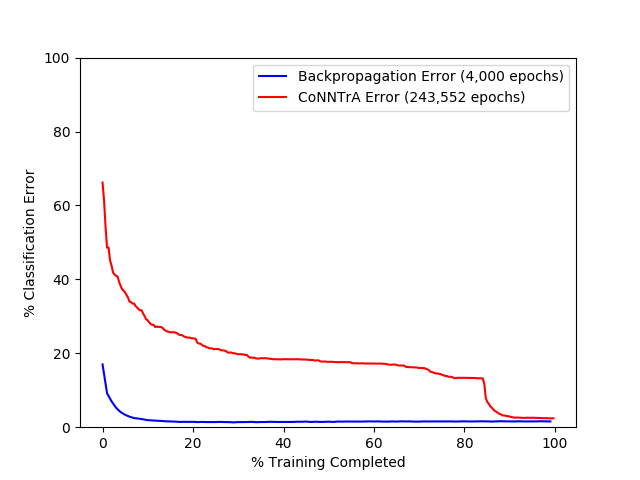}
        \caption{Validation Error Comparison (MNIST CNN)}
        \label{fig:mnist-cnn-validation-error}
    \end{subfigure}
    \caption{Error comparison for MNIST CNN models}
    \label{fig:mnist-cnn-errors}
\end{figure}

We use the LeNet architecture proposed by LeCun et al. \cite{lecun1998gradient}.
The training results for MNIST CNN models are shown in Table \ref{tab:mnist-cnn-comparison} and Figure \ref{fig:mnist-cnn-errors}.
The training and validation errors for Backpropagation are $1.40\%$ and $1.56\%$ respectively, and those for CoNNTrA are $2.60\%$ and $2.39\%$ respectively.
The memory usage for Backpropagation is $649.55$ kilobytes, while that for CoNNTrA is $20.30$ kilobytes.
While Backpropagation takes $121.97$ seconds, CoNNTrA takes $4,871.04$ seconds to complete training.
Figure \ref{fig:mnist-cnn-errors} shows the training and validation errors for Backpropagation (blue) and CoNNTrA (red).
The final training and validation errors obtained by both models are around $1-3\%$, which is the state of the art for the LeNet CNN, and corresponds to an accuracy of $97-99\%$.
The rate of convergence for CoNNTrA, shows an interesting behavior.
The rate of convergence gradually decreases until just over $80\%$ of training is completed, after which, it starts decreasing rapidly and converges to $2.60\%$ in Figure \ref{fig:mnist-cnn-training-error} and $2.39\%$ in Figure \ref{fig:mnist-cnn-validation-error}.
We attribute this behavior to the following reason.
At every training epoch in CoNNTrA, we pick a weight at random and perform a global search across all possible values to find a value that yields the smallest possible error.
When the rate of convergence started decreasing rapidly, a weight was picked which had a high impact on the classification error.
When a global search was performed for this weight, it drastically improved the error and transformed the neural network function in such a way that there was abundant room to improve the error for subsequently selected weights.

\subsubsection{Iris}


\begin{figure}[t!]
	\centering
	\includegraphics[trim={50 0 70 0}, scale=0.29]{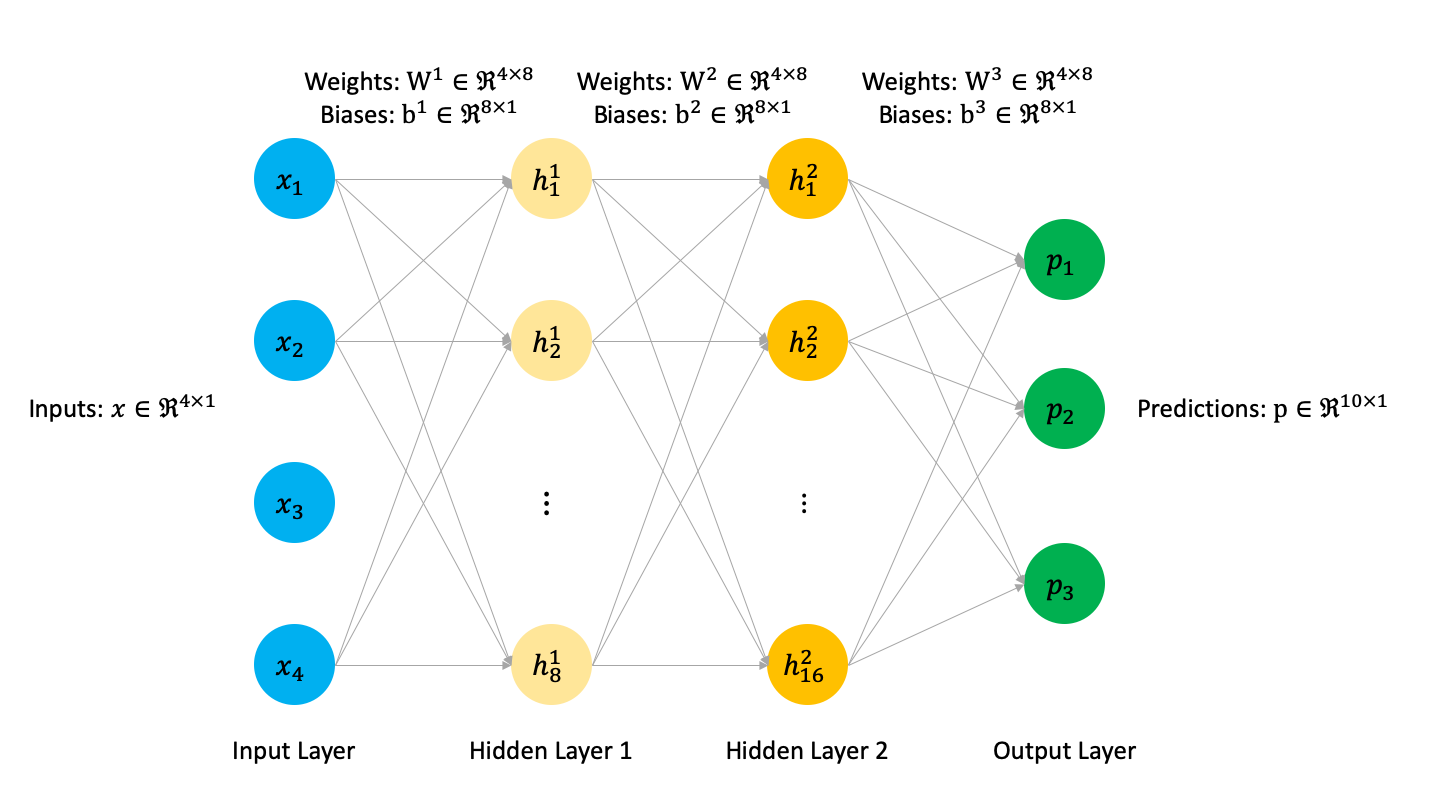}
	\caption{Schematic diagram of Iris multi-layer perceptron}
	\label{fig:iris-mlp-model}
\end{figure}

\begin{table}[t!]
    \centering
    \caption{Performance metrics for Iris DNN}
    \begin{tabular}{m{0.2\textwidth} m{0.11\textwidth} m{0.07\textwidth}}
        \noalign{\smallskip} \hline \noalign{\smallskip}
        Performance Metric & Backpropagation & CoNNTrA \\
        \noalign{\smallskip} \hline \noalign{\smallskip}
        \textbf{Training Error (\%)} 		  & \textbf{1.67}		& \textbf{1.67} \\
        \textbf{Validation Error (\%)} 		  & \textbf{3.33}		& \textbf{3.33} \\
        \textbf{Memory Usage (kilobytes)} 	  & \textbf{1.88} 		& \textbf{0.06} \\
        Training Time (seconds) 	          & 4.56 		        & 4.92 \\
        \noalign{\smallskip} \hline \noalign{\smallskip} 
    \end{tabular}
    \label{tab:iris-dnn-comparison}
\end{table}

\begin{figure}[t!]
    \centering
    \begin{subfigure}{0.4\textwidth}
        \centering 
        \includegraphics[scale=0.4]{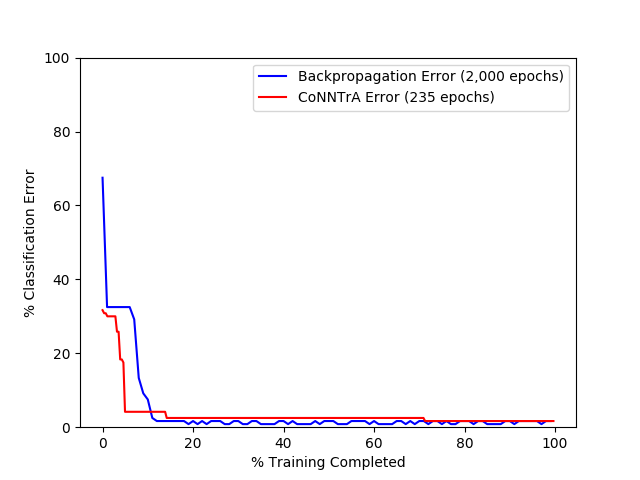}
        \caption{Training Error Comparison (Iris DNN)}
        \label{fig:iris-dnn-training-error}
    \end{subfigure}
    \begin{subfigure}{0.4\textwidth}
        \centering 
        \includegraphics[scale=0.4]{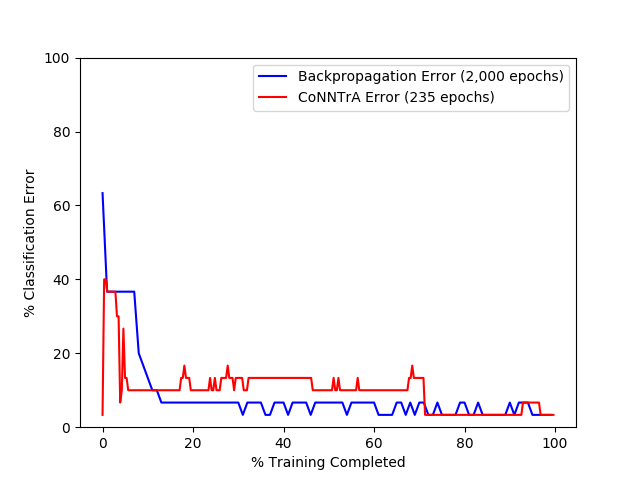}
        \caption{Validation Error Comparison (Iris DNN)}
        \label{fig:iris-dnn-validation-error}
    \end{subfigure}
    \caption{Error comparison for Iris DNN models}
    \label{fig:iris-dnn-errors}
\end{figure}

We use a three layer deep multi-layer perceptron model having two hidden layers for this classification task.
Figure \ref{fig:iris-mlp-model} shows a schematic diagram of the multi-layer perceptron model.
Each neuron in the hidden layers is indexed using a superscript and a subscript.
The superscript indicates the layer index and the subscript indicates the neuron index within that layer.

Table \ref{tab:iris-dnn-comparison} shows the performance metrics for the Iris DNN models.
Both models achieved the same training and validation errors, i.e. $1.67\%$ and $3.33\%$ respectively.
The memory usage for Backpropagation was $1.88$ kilobytes, while that for CoNNTrA was $0.06$ kilobytes.
The training time for Backpropagation was $4.56$ seconds and that for CoNNTrA was $4.92$ seconds.
Figure \ref{fig:iris-dnn-errors} shows the plot of training and validation errors for Backpropagation (in blue) and CoNNTrA (in red).
Training errors for both algorithms follow each other closely and converge at $1.67\%$.
Validation error for CoNNTrA is seen to vary abruptly initially until it starts to converge at around the $15\%$ mark with sporadic spikes.
The validation errors for both algorithms converge to $3.33\%$.

\subsubsection{ImageNet}

\begin{table}[t!]
    \centering
    \caption{Performance metrics for ImageNet CNN}
    \begin{tabular}{m{0.2\textwidth} m{0.11\textwidth} m{0.08\textwidth}}
        \noalign{\smallskip} \hline \noalign{\smallskip}
        Performance Metric & Backpropagation & CoNNTrA \\
        \noalign{\smallskip} \hline \noalign{\smallskip}
        Training Error (\%) 		        & 15.12			        & 16.98 \\
        \textbf{Validation Error (\%)} 		& \textbf{18.62}	    & \textbf{18.50} \\
        \textbf{Memory Usage (kilobytes)} 	& \textbf{499,026.75}	& \textbf{15,594.59} \\
        Training Time (seconds) 	        & 388,764.43 	        & 647,249.96 \\
        \noalign{\smallskip} \hline \noalign{\smallskip} 
    \end{tabular}
    \label{tab:imagenet-cnn-comparison}
\end{table}

\begin{figure}[t!]
    \centering
    \begin{subfigure}{0.4\textwidth}
        \centering 
        \includegraphics[scale=0.4]{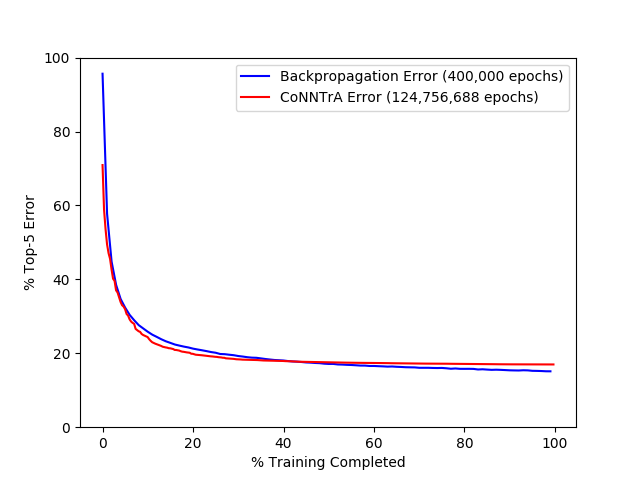}
        \caption{Training Error Comparison (ImageNet CNN)}
        \label{fig:imagenet-cnn-training-error}
    \end{subfigure}
    \begin{subfigure}{0.4\textwidth}
        \centering 
        \includegraphics[scale=0.4]{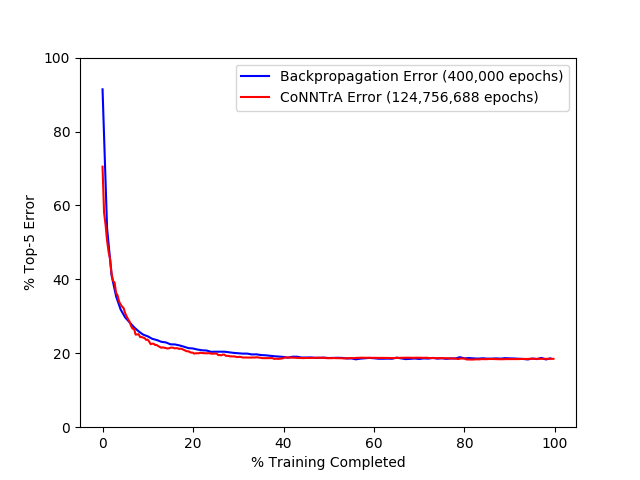}
        \caption{Validation Error Comparison (ImageNet CNN)}
        \label{fig:imagenet-cnn-validation-error}
    \end{subfigure}
    \caption{Error comparison for ImageNet CNN models}
    \label{fig:imagenet-cnn-errors}
\end{figure}

We use the AlexNet CNN proposed by Krizhevsky et al. \cite{krizhevsky2012imagenet}.
Table \ref{tab:imagenet-cnn-comparison} shows the performance metrics.
While Backpropagation takes $388,764.42$ seconds, CoNNTrA takes $647,249.96$ seconds to complete training.
The training errors for Backpropagation and CoNNTrA are $15.12\%$ and $16.98\%$ respectively, and the validation errors are $18.62\%$ and $18.50\%$ respectively.
All errors computed for ImageNet CNN model are top-5 errors.
The memory used by Backpropagation is $499,026.75$ kilobytes, and that used by CoNNTrA is $15,594.59$ kilobytes.
Figure \ref{fig:imagenet-cnn-errors} shows the training and validation errors for training the ImageNet CNN model using Backpropagation (blue) and CoNNTrA (red).
We see a regular trend in Figure \ref{fig:imagenet-cnn-errors}.
Both models converge to around $15-17\%$ training error and $16-18\%$ validation error, which are in the same ballpark as the state of the art errors for the AlexNet model.

\subsection{Discussion}

\begin{figure}
	\centering
	\begin{subfigure}{0.32\textwidth}
		\centering
		\includegraphics[scale=0.32]{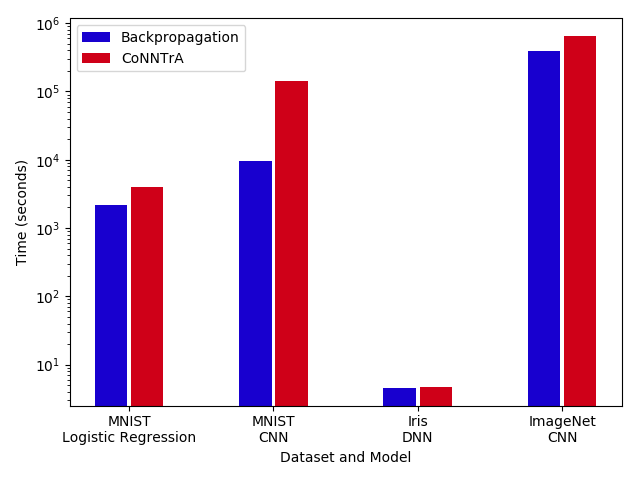}
		\caption{Training Time}
		\label{fig:training-time-bar}
	\end{subfigure}
	\\
	\begin{subfigure}{0.32\textwidth}
		\centering
		\includegraphics[scale=0.32]{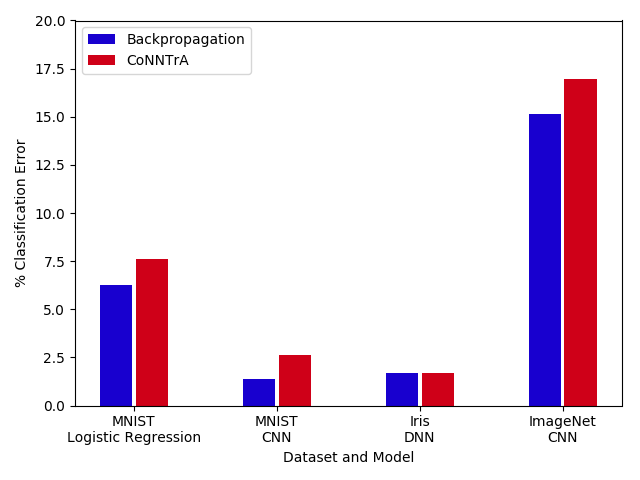}
		\caption{Training Error}
		\label{fig:training-error-bar}
	\end{subfigure}
	\\
	\begin{subfigure}{0.32\textwidth}
		\centering
		\includegraphics[scale=0.32]{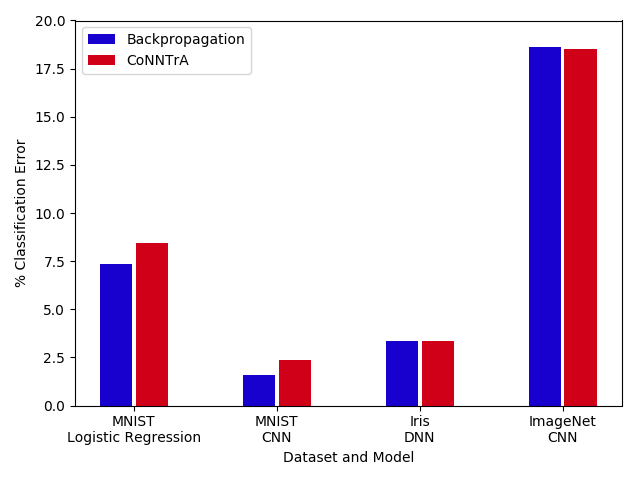}
		\caption{Validation Error}
		\label{fig:validation-error-bar}
	\end{subfigure}
	\\
	\begin{subfigure}{0.32\textwidth}
		\centering
		\includegraphics[scale=0.32]{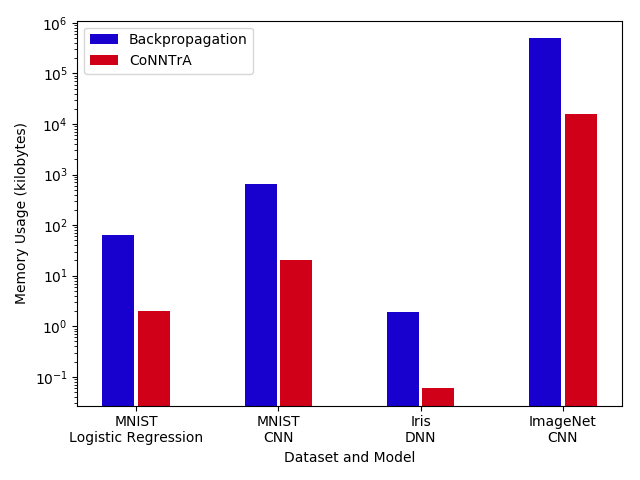}
		\caption{Memory Usage}
		\label{fig:memory-bar}
	\end{subfigure}
	\caption{Comparison of Backpropagation and CoNNTrA}
	\label{fig:consolidated-results}
\end{figure}

Figure \ref{fig:consolidated-results} presents the performance of Backpropagation and CoNNTrA in a consolidated fashion.
The X-axis shows datasets and models, and the Y-axis shows the performance metric.
Blue and red bars denote Backpropagation and CoNNTrA performance metrics respectively.
In Figure \ref{fig:training-time-bar}, we observe that CoNNTrA takes more time to complete training for all tasks.
This is because we were using a serialized implementation of CoNNTrA as our objective was to demonstrate a proof of concept that CoNNTrA is able to train models having accuracies at par with Backpropagation.
With parallel implementations of CoNNTrA, we expect the training times to significantly reduce.

In Figures \ref{fig:training-error-bar} and \ref{fig:validation-error-bar}, the training and validation errors for Backpropagation and CoNNTrA are within the same ballpark and are at par with the state of the art error values for the respective models.
The CoNNTrA errors fall slightly short of Backpropagation errors for all models except Iris DNN.
Given the same architecture of the models for both Backpropagation and CoNNTrA, the CoNNTrA weights were ternary whereas Backpropagation used double precision floating point weights.
Going from double precision to ternary weights resulted in less expressibility for CoNNTrA, because of which we see slightly higher value of training and validation errors.

In Figure \ref{fig:memory-bar}, the memory usage for CoNNTrA is about $32 \times$ less than Backpropagation for all models.
It requires $2$ bits to store each ternary valued CoNNTrA weight, and $64$ bits to store each double precision floating point Backpropagation weight.
A $32\times$ reduction in memory usage is extremely significant in edge computing applications, especially embedded systems, Internet of Things, autonomous vehicles etc.

\section{Conclusion}
\label{sec:conclusion}
Edge computing systems in applications like Internet of Things (IoT), autonomous vehicles and embedded systems in the post Moore's law era will require machine learning models that not only produce low error and train fast, but also consume low memory and power.
While traditional learning algorithms like Backpropagation can train deep learning models in a reasonable amount of time and obtain low error, they consume significantly large memory and power.
In this work, we propose a novel learning algorithm called Combinatorial Neural Network Training Algorithm (CoNNTrA), which is a coordinate gradient descent-based algorithm that can train deep learning models having constrained learning parameters, for example, having binary or ternary values.

The objective of this study was to demonstrate that CoNNTrA can train deep learning models with constrained learning parameters, which yield errors at par with the Backpropagation models, \emph{and} consume significantly less memory.
We presented CoNNTrA in Section \ref{sec:conntra} along with its theoretical underpinnings and complexity analysis.
In Section \ref{sec:performance-evaluation}, we used CoNNTrA to train deep learning models for three machine learning benchmark data sets (MNIST, Iris and ImageNet).
We demonstrated that CoNNTrA can train these models having errors in the same ballpark as Backpropagation models.
More importantly, we showed that the CoNNTrA models consume $32\times$ less memory than the Backpropagation models.

In our future work, we would like to implement CoNNTrA in an efficient parallelized fashion to improve the training times.
We believe that such a parallel implementation of CoNNTrA would be able to train deep learning models that are not just accurate and consume orders of magnitude less memory than Backpropagation, but can also be trained efficiently.
This would be invaluable to training machine learning and deep learning models in the post Moore's law era, especially for edge computing systems supporting critical applications.
We would also like to study the applicability of CoNNTrA for solving other NP-complete problems like traveling salesman problem, protein folding, and genetic imputation.

\bibliographystyle{IEEEtran}
\bibliography{reference}

\end{document}